%% file: main.tex
\documentclass[10pt,twocolumn,letterpaper]{article}

\usepackage{iccv}
\usepackage{times}
\usepackage{epsfig}
\usepackage{graphicx}
\usepackage{subcaption}
\usepackage{amsmath}
\usepackage{amssymb}
\usepackage[table,xcdraw,dvipsnames]{xcolor}
\usepackage{makecell}
\usepackage{multirow}

\usepackage[pagebackref=true,breaklinks=true,letterpaper=true,colorlinks,bookmarks=false]{hyperref}

\iccvfinalcopy

\ificcvfinal\fi

\definecolor{mydarkblue}{rgb}{0,0.1,0.6}
\definecolor{Gray}{gray}{0.9}
\definecolor{Goldenrod}{RGB}{245,245,220} %

\newcommand{\markcell}[1]{\cellcolor{Gray!36}\textbf{#1}}
\newcommand{\dataname}{{OxfordTVG-HIC (Humorous Image Captions)}}
\newcommand{\dname}{{OxfordTVG-HIC}}

\newcommand{\tablestyle}[2]{\setlength{\tabcolsep}{#1}\renewcommand{\arraystretch}{#2}\centering\small}

\hypersetup{
    colorlinks=true,
    citecolor=RoyalBlue}

\begin{document}

\title{\dname: Can Machine Make Humorous Captions from Images?}

\author{
      Runjia Li$^{1}$\footnotemark[1]
	 \;\; Shuyang Sun$^{1}$\footnotemark[1]
	\;\; Mohamed Elhoseiny$^2$
	\;\; Philip Torr$^1$\\
	$^1$Torr Vision Group, University of Oxford \;\; $^2$KAUST\\
 \url{https://torrvision.com/tvghic/}
}

\maketitle
\ificcvfinal\thispagestyle{empty}\fi

\renewcommand{\thefootnote}{\fnsymbol{footnote}}
\footnotetext[1]{These authors contributed equally to this work. Correspondence to Shuyang Sun (\url{kevinsun@robots.ox.ac.uk}). }

\begin{abstract}
This paper presents \dataname, a large-scale dataset for humour generation and understanding. Humour is an abstract, subjective, and context-dependent cognitive construct involving several cognitive factors, making it a challenging task to generate and interpret. Hence, humour generation and understanding can serve as a new task for evaluating the ability of deep-learning methods to process abstract and subjective information.
Due to the scarcity of data, humour-related generation tasks such as captioning remain under-explored.
To address this gap, \dname~offers approximately 2.9M image-text pairs with humour scores to train a generalizable humour captioning model.
Contrary to existing captioning datasets, \dname~features a wide range of emotional and semantic diversity resulting in out-of-context examples that are particularly conducive to generating humour. Moreover, \dname~is curated devoid of offensive content.
We also show how \dname~can be leveraged for evaluating the humour of a generated text. 
Through explainability analysis of the trained models, we identify the visual and linguistic cues influential for evoking humour prediction (and generation). We observe qualitatively that these cues are aligned with the benign violation theory of humour in cognitive psychology.

\end{abstract}

\input{figures/teaser}

\section{Introduction}
Humour has been recorded as a universal and high-level cognitive perception since Sumerians wrote down the first joke and remains a complicated concept due to its dependence on culture, visual and linguistic stimuli, as well as fundamental affective factors. Generating humorous content poses a significant challenge. As the arousal-reduction mechanism~\cite{least_effort} suggests, the optimal level of novelty for humour should be neither too low nor too high, so that most audience can comprehend and appreciate it. Achieving this balance can be difficult.
   Another study from neuroscience~\cite{neuroscience_humor} states: ``successful jokes involve a cognitive juxtaposition of mental sets, followed by an affective feeling of amusement''. Linguistically, semantic and phonological violations of mental sets interact to generate humour~\cite{benign_violation}. When multiple modalities such as vision and language are involved, the complexity of juxtaposition will significantly increase. Thus, Humour captioning, as a multi-modal humour generation task, can be a useful task to investigate the upper limit of deep learning to handle high-level abstraction and creativity. 
   
  The definition of humour captioning is not clear and the task is under-explored. We argue that humour captioning should be distinguished from conventional image captioning ~\cite{mscoco, conceptual, flicker}, which mainly describes the objects or scenes in the image. The main goal of humour captioning should be to elicit a cognitive perception of humour rather than to provide factual information. This implies that a humorous sentence may not be relevant to the image content and still be a valid example. Hence, standard metrics~\cite{bleu, meteor, cider, rouge} that assess the quality of captions and conventional image captioning training framework may not be appropriate for humour captioning. 

Humour captioning has received little attention. An initial attempt~\cite{dank_learning} used a relatively simple model~\cite{show_and_tell} and a small-scale dataset and followed the conventional image captioning framework. Another recent image-text dataset~\cite{openai_humor} that aims to evaluate how well deep-learning models can comprehend humour has too few images (1.6k) with limited diversity to train a robust model.

To overcome the limitation faced by previous research, we introduce \dataname, a large-scale image-text dataset for humour captioning. To our knowledge, it is the first dataset of this kind that contains 2.9 million image-text pairs with a humour score to measure their funniness. In \dname, each image has 53.7 captions on average, while the conventional image caption datasets (e.g., ~\cite{mscoco, conceptual, flicker})  have less than 5 captions per image.
\input{tables/table_caption_dataset_statistics}
\input{figures/data_stats}
Based on \dname, we develop humour generators that can automatically produce humorous captions given any images. the humour generators are trained on our proposed position-conditioned loss to address the issues of low diversity and humour caused by using cross-entropy loss on \dname. 
Specifically, since \dname~is a dataset with high grammatical and semantic diversity, the cross-entropy loss tends to generate captions that are close to all ground truths given an input image, which is detrimental for training. The reason is that a generated sentence that is close to a set of distinct sentences is also far from those sentences. Therefore, the humour level and diversity are reduced with cross-entropy loss. Our position-conditioned loss solves this problem with promising results.

The qualitative nature of human thoughts makes it hard to evaluate and understand humour quantitatively. However, there are still some theories~\cite{benign_violation, relief, superior} proposed by psychologists to explain and measure humour.
Drawing insights from one of the theories: Benign violation~\cite{benign_violation}, we propose the benign violation humour score based on a humour classifier trained on \dname~ and a benign level classifier. Together with other linguistic metrics such as diversity and fluency, we benchmark the performance of our model on \dname. 

Moreover, the learned ability to perceive humour as digital signals sheds light on the quantitative understanding and explainability of humour. From our observation and analysis, the model regards the abnormal and emotionally intense parts of images and sentences as the most important factors for creating humour. We hypothesise our observation is consistent with the Benign violation theory~\cite{benign_violation}.

In summary, \textbf{our main contributions} are as follows:
     
     1. We introduce \textit{\dname}, a large-scale humour-oriented image-text dataset that addresses the lack of data in image-text-based humour generation and detection. 
     
     2. We show the diversity and richness of \textit{\dname}~and design the position-conditioned loss to better train a humour caption generator on the diverse data of \textit{\dname}.
     
     3. By explaining the learned humour models, we analyse the visual and linguistic stimuli that evoke humour and provide insights regarding humorous cues within the data. We further observe that the insights are aligned with the Benign violation theory~\cite{benign_violation}.

\section{Background and related work}

\paragraph{Humour in deep learning.}
Humour is a complex cognitive phenomenon that requires a high level of abstraction to comprehend. Even humans may struggle to articulate the mechanisms and sources of humour when we encounter it. Having a sense of humour is so hard and valuable for humans that we have professional comedians to make jokes. Nowadays deep learning models have achieved remarkable performance in object-centric multi-modal tasks, but their ability to generalise to tasks involving abstract and high-level concepts like humour remains uncertain. Previous research has attempted to extract and analyse abstract concepts from various modalities, especially humour detection in texts~\cite{16000oneliner, ted_laughter, ptt_jokes, pun_of_day, rjokes, colbert} and videos~\cite{funnynet, laughing_machine, masm, hkt}, as well as text-based joke generation~\cite{joking_bird, pun_gan, pun_generation,humor_headline}. 

\noindent\textbf{Image captioning data.}
The goal of image captioning is to generate natural language descriptions for a given image. Most existing datasets~\cite{mscoco, conceptual, flicker} focus on the physical aspects of the scenes in an image and evaluate the descriptive skills of the models. Some recent datasets~\cite{artemis, artelingo} introduce the challenge of capturing the emotional effects of the visual stimulus and expressing them in natural language. However, to the best of our knowledge, no public large-scale dataset exists that specifically targets humour generation in captioning and examines how visual and linguistic cues trigger humour as a high-level cognitive phenomenon.

\noindent\textbf{Humour-oriented caption generation.}
Despite its potential applications and benefits, humour-oriented captioning remains a relatively under-explored area. An initial attempt~\cite{dank_learning} adapted a simple model~\cite{show_and_tell} to a small self-collected dataset and another  dataset~\cite{openai_humor} was proposed to benchmark the machine’s ability to comprehend humour. However, these datasets are limited in size and diversity and do not support the development of a robust model. Thus, a new large-scale humour captioning dataset is in need. Moreover, existing approaches have not tailored their image captioning modules to the specific features of humour-oriented captioning and have relied on conventional metrics that may not capture the humorous aspects of the captions. Though humour is complex to measure, psychologists have proposed several criteria such as Benign violation theory~\cite{benign_violation}, Relief theory~\cite{relief}, and Superiority theory~\cite{superior} to measure humour. Rather than using standard metrics~\cite{bleu, cider, rouge, meteor}, redesigning evaluation metrics for humour captioning based on the psychological theories of humour should be more appropriate.

\section{Humour Dataset: \dname}\label{sec:dataset}

Current public datasets~\cite{dank_learning, openai_humor} for humour-oriented image captioning contain too few image samples for the trained model to generalise well on unseen images. The lack of a large-scale available dataset hinders the development and generalisation of humour-centric multi-modal tasks such as humour captioning.
\input{tables/table_emotion_div}
Therefore, we present a large-scale humour captioning dataset: \dname, containing \textbf{2,885,326} image-text pairs in English and \textbf{53,728} unique images acquired from popular online websites and communities including Bokete Ogiri (Japanese)~\cite{bokete}, ImgFlip (English)~\cite{imgflip},  and online image generators (English)~\cite{meme_generator}. Each image-text pair is annotated with a funny score in the form of votes from the website users, which could be crucial to quantitatively evaluate how much humour the visual and linguistic stimuli create jointly. 

All raw captions in \dname~are translated into English with DeepL translator~\cite{deepl} and cleaned with MangaOCR~\cite{manga_ocr} and TextBlob~\cite{textblob} to ensure no linguistic hints appear in the images and captions with non-English words caused by homophonic translation are deleted so that culturally-exclusive contents and words will be removed.

\input{tables/table_grammar}
\subsection{Data statistics}
Rather than attending to the physical content of an image, humour captioning focuses on a  perceived aspect of the image and further develops a thought that relates to the objects, the objects' point of view, or the audience's comments on the image. The diverse form of humour indicates that the distribution of appropriate humour captions for an image can be much wider than object-centric captions. More specifically, the \dname~stands its way out due to the fact that humour can take a variety of forms, or in other words, an infinite number of semantically distinct captions can stimulate humour in combination with the image. Another significant difference is that \dname~has more emotional richness than object-oriented image captioning datasets, which requires more ability to extract abstract and psychologically high-level concepts from the input image. For this reason, as shown in Figure \ref{fig:dataset_statistics}, \dname~has 53.7 captions per image on average and most of the images from the \dname~have 10 to 1,000 captions, which is significantly higher than that of popular image captioning datasets~\cite{mscoco, conceptual, flicker, artemis, artelingo} as demonstrated in Table \ref{tab:caption_dataset_statistics} and Figure \ref{fig:dataset_statistics}. The high number of captions per image poses a greater challenge for captioning models to tackle than conventional image captioning.

\subsection{Richness and diversity}
Unlike object-centric image captioning datasets~\cite{mscoco, conceptual, flicker} that emphasise the physical relationship between objects in the image, \dname~captures high-level mental perception in its captions. The subjectivity of \dname~results in a rich and diverse set of emotions, semantics, and grammar.

\noindent\textbf{Emotional richness.}
Though not an emotion itself, humour is built on emotions~\cite{neuroscience_humor}. Quantitatively reflected in the dataset statistics, $63.7\%$ captions in the \dname~are detected to be emotional by DistilRoBERTa~\cite{hartmann2022emotionenglish}, compared to only $39.1\%$ captions in COCO~\cite{mscoco} are considered emotional as shown in Table \ref{tab:emotion_distribution}. In more detail, when analysed apart from the image, the captions in \dname~have more emotion in surprise, disgust, and anger than the object-centric dataset. Thus, although humour is generally considered a positive feeling, its emotional components could be non-positive.

\input{figures/figure_data_semantic_dist}
\noindent\textbf{Grammar pattern diversity.}
 As shown in Table \ref{tab:grammar_diversity}, if we consider the relative position of the nouns, verbs, adjectives, and conjunctions as grammar patterns, the captions in \dname~have significantly more grammar types than other popular image captioning datasets~\cite{mscoco, conceptual, flicker, artemis, artelingo}. The grammatical diversity results from the fact that  humour can take an infinite number of forms due to its subjectivity. 

\noindent\textbf{Semantic diversity.} 
The semantic diversity of humour captions is another characteristic of \dname. The captions for the same image in \dname~may have different meanings when considered in isolation. To illustrate this, we used the CLIP~\cite{clip} text encoder to obtain sentence-level semantic embeddings as shown in Figure \ref{fig:data_set_semantics}. In COCO~\cite{mscoco}, the semantic embeddings of captions for the same image tend to form a cluster, indicating that they have similar meanings. However, in \dname~, the semantic embeddings of captions for the same image are scattered across the embedding space. Moreover, we observed that some semantic embeddings of captions from different images are close to each other in the embedding space. This suggests that semantically similar captions can generate humour for different images.
\input{figures/figure_position}

\subsection{Ethics}
The raw dataset contained some offensive content such as profanity, sexual references, and hate speech based on race, gender, or belief. To ensure the quality and safety of \dname, we applied several filters to remove the harmful content at different levels: image-level: NSFW offensive content detection~\cite{nsfw}, text-level: filtering captions with words from English profane word dictionary, and image-text-level: Villo~\cite{villo} hateful memes detection. Finally, we discarded 10,352 images and 133,341 captions from the raw dataset.

\section{Humour Captioning Baseline}
Humour captioning is different from conventional image captioning~\cite{mscoco, conceptual, flicker}, which aims to describe the objects or scenes in the image. The primary objective of humour captioning is to evoke the perception of humour rather than to convey factual information. This means that a humorous sentence can be valid even if it is not relevant to the physical image content, which introduces high grammatical and semantic diversity into the ground truths in \dname~as shown in Figure \ref{fig:dataset_demo}.  In this paper, the diversity in grammar and semantics is taken into account by our baseline humour generator. 

\subsection{Humour Generator}

\input{figures/figure_generator_demo}
We experiment with two popular backbone architectures when designing humour generators trained on \dname: ClipCap~\cite{clipcap}, which combines CLIP~\cite{clip} image encoder with a GPT-2~\cite{gpt2} text decoder connected by a mapping network; and the recent captioning model BLIP~\cite{li2022blip}, which utilises the noisy web data by bootstrapping the captions. Those two backbones are trained with a position-conditioned loss we designed specifically for \dname.

\subsection{Position-conditioned Loss.} \label{sec:position loss}
Token-wise cross-entropy is a commonly used supervision loss in natural language processing tasks, and prominent language models such as BERT~\cite{bert}, GPT2~\cite{gpt2}, and RoBERTa~\cite{roberta} have been trained using this method. While cross-entropy loss has demonstrated effectiveness in object-centric image captioning tasks, it may introduce confusion to models during training when presented with diverse captions for the same image that differ significantly in semantics and grammar patterns. This confusion is particularly troublesome when predicting the first few tokens of a sentence, as the initial label tokens for the same image may vary significantly in meaning as demonstrated in Figure \ref{fig:position}. Training the model using cross-entropy loss under these circumstances leads to the model predicting the most similar token to all different label tokens to minimise the loss. However, if the predicted token is comparable to all label tokens that are distinct from one another, it will also be dissimilar to all of those label tokens, resulting in generated tokens that lack humour and are "neither fish nor fowl" as illustrated in Figure \ref{fig:generator_demo}. 

To tackle the problem, we propose a token-position-conditioned loss based on the assumption that the closer to the start of the caption, the less strict the condition is for the predicted token. For example, the condition of the first token on the image prompt should be weak because the same image prompt could lead to a variety of distinct first tokens. In more detail, we do not wish to penalise false negative prediction too much at the beginning of the sentence because the wrong prediction for one label could potentially be correct for another. For implementation, we designed a false positive loss weight function $w_j$ based on token position ($j$ is the $j$-th token position in the caption). Mathematically, the position-conditioned loss $\mathcal{L}_j$ can be formulated as the following:
\begin{equation}
    \mathcal{L}_j = -\frac{1}{B}\sum_{i=1}^B\sum_{z=1}^K     \begin{cases}
     \log({c}_{ij}^{z}) & {y}_{ij}^{z} = 1  \\
      w_j\log(1-{c}_{ij}^{z}) & {y}_{ij}^{z} = 0\\ 
    \end{cases}
\end{equation}
where $\mathbf{y}_{ij} = [y_{ij}^{1}, ..., y_{ij}^{K}] \in \{0, 1\}^{K}$ is a one-hot vector for the label token and $c_{ij}^{z}$ represents the $z$-th word probability of the predicted token $\mathbf{c}_{ij} \in [0, 1]^{K}$, and $K$ is the number of word tokens in the label space. $B$ represents the data batch.

The false positive loss weight function $w_j$ could take various forms. One general principle for choosing $w_j$ is that the function needs to be monotonically increasing as the position index increases because we assume the condition gets stricter at the end of the caption. Next, we discuss choices for the false negative loss weight function $w_j$.

\noindent\textbf{Linear.} The most straightforward monotonically increasing function is the linear function. In this paper, we consider:
\begin{equation}
    w_j = \gamma\frac{j}{M}, w_j \in \mathbb{R}^+
\end{equation}
where $M$ denotes the number of tokens in the caption and $\gamma$ is a hyperparameter that controls how drastically the false positive penalty should increase as the token position moves along the sentence. 

\noindent\textbf{Gaussian.}
Gaussian distribution in the right half plane has properties of smooth change and monotonically decreasing at a fixed range. If we consider the transformed Gaussian function described as the following:
\begin{equation}
    w_j = 1 - e^{-({\frac{j}{\beta}})^2}, w_j \in \mathbb{R}^+
\end{equation}
where $\beta$ controls the standard deviation, then the false negative loss weight function will be monotonically increasing in the right half plane and ranges from $[0, 1]$.

\noindent\textbf{Sigmoid.}
The sigmoid function is another choice of monotonically increasing function with nice properties to smoothly converge at a constant value as the position index increases. We consider the sigmoid weight function as:

\begin{equation}
    w_j = \frac{2}{1+e^{-\frac{j}{\alpha}}}-1, w_j \in \mathbb{R}^+
    \label{eq:sigmoid}
\end{equation}
where $\alpha$ is a hyper-parameter that controls the scaling of the sigmoid function, or in other words, how fast the false negative loss weight increases. In our ablation experiment, we find $\alpha=6$ gives the highest humour intensity.

\section{Evaluation}

Object-centric image captioning task uses linguistic evaluation metrics such as BLEU~\cite{bleu}, ROUGE~\cite{rouge}, METEOR~\cite{meteor}, and CIDEr~\cite{cider} to measure how close the predicted caption is to the ground truth objectively. And all previous work~\cite{dank_learning} on humorous caption generation used the objective metrics above to evaluate their humour generation model. However, we believe those linguistic metrics are inappropriate for the humour-centric image captioning task where ground truths per image could be distinct from one another in both semantics and grammar patterns.

To evaluate humour, we borrow insights from psychology and propose a new evaluation metric: Humour score and Benign score, which we believe serve better as a proxy for humour intensity evaluation to our task than conventional linguistic metrics mentioned above.

\subsection{Humour metrics}
Humour has been studied for decades, some famous explanations include the Benign violation theory~\cite{benign_violation}, Superiority theory~\cite{superior}, and Relief theory~\cite{relief}. The Benign violation theory~\cite{benign_violation} states that humour happens when a statement violates how the audience thought the world ought to be and the violation sounds benign to the audience. The subject of violation includes social norms, linguistic norms, moral norms, and self-dignity. An example of benign violations is shown in Figure \ref{fig:dataset_demo}. We believe designing an evaluation metric based on Benign violation theory~\cite{benign_violation} could allow us to measure how well a model can generate humour.

\noindent\textbf{Humour Score.}
In this paper, the approach for evaluating the humour intensity in image-text pairs is based on a classification model trained on \dname. Positive samples are drawn from \dname, while negative samples of an image are comprised of semantically distinct captions from other images, randomly generated text from GPT-3~\cite{gpt3}, and generated captions from captioning models trained on COCO~\cite{mscoco}. The model is trained to distinguish between these positive and negative samples and ultimately to output the humour confidence given an image-text pair, which is positively correlated with the violation level. 
The classifier is composed of 3 components: ResNet50~\cite{resnet} image encoder, GPT-2~\cite{gpt2} text encoder, and linear classification head to conduct a binary classification task. The classifier is trained on binary cross entropy loss and achieved an in-domain accuracy of 89\% and out-domain accuracy of 77\% (in-domain means seen images; out-domain means unseen images). The relative non-perfect performance of the evaluator implies the difficulty to comprehend and judge humour.

\noindent\textbf{Benign Score.}
According to the benign violation theory, to generate humour, the benign level should be high and no offensive content to the audience should be included. We measure the benign value by Villo~\cite{villo}, a classification model trained on the Hateful Memes Challenge dataset~\cite{hateful_memes}. Given an image-text pair, the model can tell whether the information the image-text pair conveys is hateful or offensive. The benign score of an image-text pair is the output probability of the Villo~\cite{villo} model.  

The Benign score and Humour scores are positively correlated with the humour intensity of the generated captions.

\subsection{Linguistic metrics}
Besides the humour evaluation, fluency and diversity should also be evaluated to measure how well the generated sentences make sense and diversify linguistically. 

\noindent\textbf{Fluency score.}
We use Parrot~\cite{parrot}, a T5-based language model, to measure the fluency score of the generated sentences. Parrot was originally designed to paraphrase a given sentence and also includes an evaluation model to measure the fluency of its paraphrased sentence. We argue that Parrot is suitable as a fluency evaluation model for the image captioning tasks because Parrot’s original purpose aligns with our goal to evaluate generated sentences.

\noindent\textbf{Diversity score.}
CLIP~\cite{clip} is a powerful pre-trained model to extract semantic information from images and texts. We use CLIP~\cite{clip} as a semantic diversity evaluator based on the cosine similarity of the extracted semantic features from generated captions. Mathematically, let $p$ represent the caption semantic feature extracted by CLIP~\cite{clip}. Then the diversity score can be described as :
\begin{equation}
    diversity = \sqrt{1 - (\frac{1}{N}\sum_{m=1}^{N} \max_{n, n\neq m} \frac{p_m^Tp_n}{|p_m||p_n|})^2}
\end{equation}

\section{Experiments}

\subsection{Humour caption generation}

\noindent\textbf{Implementation details.}
We train all models on \dname~with COCO~\cite{mscoco} pre-trained weights. The training process consists of 20 epochs with 500 warm-up steps for both cross-entropy and position-conditioned loss. We use the cosine learning rate decaying schedule with the initial learning rate of $10^{-5}$. 
\input{tables/table_humor_level}

\input{tables/table_kernel_function}
\noindent\textbf{Experimental results.}
As shown in Table \ref{tab:humor level}, humour generators trained on \dname~obtain at most $56.2\%$ more in humour score than those trained on COCO~\cite{mscoco}, and humour generators trained with position-conditioned loss improve humour score by $18.2\%$ compared to those trained with cross-entropy. The position-conditioned loss improves the diversity by $13\%$ (as shown in Table \ref{tab:ablation}) with comparable fluency.  

\subsection{Ablation study}

\paragraph{Influence of different kernel functions.}

The sigmoid function of the position-conditioned loss conveys the highest benign and humour score as well as diversity as demonstrated in Table \ref{tab:kernel_function}. 

\noindent\textbf{Influence of kernel function hyperparameters.} The hyper-parameter $\alpha$ for the sigmoid kernel function of the position-conditioned loss controls how fast the false positive penalty increases with positions. The motivation to introduce the position-conditioned loss is to overcome the problem of limited diversity caused by cross-entropy and $\alpha$ plays an important role in diversity control. As demonstrated in Table \ref{tab:ablation}, when $\alpha$ increases, the diversity of generated captions increases as well while the humour level seems without correlation.
\input{tables/table_ablation}
\input{figures/figure_attentionmap_demo}
\input{figures/figure_grad_cam_demo}
\subsection{Understanding humour}
The humour generator and classifier can both provide insight into the quantitative understanding of humour, which is one of the major goals of this paper. In general, we aim to quantify what visual and linguistic clues the models use to generate and evaluate humour. 

\noindent\textbf{Generator attention visualisation.} We visualise~\cite{attention_visual} the attention map of the image encoder and examine how it attends to different parts of the image to produce humorous captions. Figure \ref{fig:attention_demo} shows the comparison of the attention heatmaps between the image encoder trained on COCO~\cite{mscoco} and \dname. The results indicate that the \dname-trained encoder pays more attention to affective features in the image. For example, it focuses more on the facial expressions of people than the COCO~\cite{mscoco}-trained encoder, as they often convey rich emotions. Moreover, it highlights some unusual or exaggerated aspects of facial expressions, such as a dog’s odd smile or a child’s evil eyes. These features suggest that humour arises from some form of incongruity or violation of mental sets. This observation is consistent with the benign violation theory~\cite{benign_violation} of humour. 

\noindent\textbf{Evaluator gradient visualisation.}
To visualise the most contributive parts of the image and text to humour generation judged by the humour classifier, we apply GradCAM~\cite{grad_cam} on images and gradient magnitude analysis on text embedding space. Figure \ref{fig:grad_cam_demo} demonstrates that the violation evaluator also focuses on facial expressions and incongruous elements in the image (\textit{e.g.} “a cat’s perplexed face” and “a car in the air”), similar to the image encoder of the generator. For the captions, pronouns seem to play a significant role as they involve the audience in the situation portrayed in the picture. This involvement and empathy may be the key to eliciting humour in the audience’s mind.
\section{Discussion and Future Works}
Besides humour-oriented image captioning, \dname~as an image-text dataset could be applied in any humour-oriented task involving visual and linguistic modalities. Potential directions include humorous image generation guided by texts and simultaneous humour-oriented image-text generation. Since \dname~is collected from different cultural websites and communities, the explainability and understanding of the cultural and linguistic differences in humour could be explored with \dname. If similar video-based datasets are created, the humour generation and evaluation tasks can be extended to a more complicated and challenging level.
\section{Conclusion}
Humour is a unique and iconic characteristic of the human cognitive process. It reflects out-of-context and unexpected interpretations of information. In this work, we pave the way for achieving humour understanding and generation of artificially intelligent systems by (1) the release of the \dname~dataset that for use in humour-oriented vision language generation and evaluation; and (2) a demonstration of humour generators that can create jokes given any images. The capacity to computationally deal with humorous attributes of text and images opens an exciting new direction in human-computer communication and interaction and unlocks a novel hallmark of cognition.

\noindent\textbf{Acknowledgement.} We would like to thank Ashkan Khakzar, Yangchen Pan, and Jindong Gu for their helpful suggestions and feedback on the paper. This work is supported by the UKRI grant: Turing AI Fellowship EP/W002981/1 and EPSRC/MURI grant: EP/N019474/1. We would also like to thank the Royal Academy of Engineering and FiveAI.

\section{Appendix}
\input{supp/table/table_hic_demo}

\input{supp/table/table_coco_demo}
\input{supp/table/table_hic_failure}

\subsection{More generated humorous captions on unseen images}
As demonstrated in Figure \ref{fig:hic_generated_demo}, the humour generators that we have trained are capable of producing humorous captions for a diverse range of unseen images from \dname. However, the perceived humour intensity of the audience may differ depending on various factors such as culture, language, and personal experience, since humour is a complex cognitive phenomenon that requires high-level and subjective understanding.

Since humour generation is such an abstract and complicated task, our humour generators sometimes fail to induce the perception of humour even while the captions are somehow related to the input image as shown in Figure \ref{fig:hic_generated_failure_demo}. 

In addition to \dname, it is our intention to evaluate the efficacy of our trained humour generators on alternative image captioning datasets. As demonstrated in Figure \ref{fig:coco_generated_demo}, the humour generators are capable of generating humorous captions for images sourced from COCO~\cite{mscoco}. From this, we can deduce that the humour generators exhibit both robustness and generalizability in the humour-oriented captioning task.

\subsection{Limitations} 
As the first large-scale study of humour captioning, we acknowledge the following limitations of our work. (1) Our humour classifier is trained on positive samples from \dname, which may bias the classifier towards generators trained on the same dataset. (2) The humour classifier only correlates to the benign violation theory, not directly linked. A stronger connection to psychological theory is desirable. We leave this for future work. (3) Because the funny scores in \dname are collected based on user votes from different humour-oriented communities, the funny scores are not entirely proportional to the true humour intensity, but only positively correlated. (4) Due to the fact that certain humorous captions are originally composed in Japanese and subsequently translated into English via DeepL~\cite{deepl}, it is possible that the intended humour of the original Japanese captions may be diminished as a result of inaccuracies in translation

{\small
\bibliographystyle{ieee_fullname}
\bibliography{egbib}
}

\end{document}

%% file: figures/teaser.tex
\begin{figure}[t]
\resizebox{\columnwidth}{!}{%
\includegraphics[width=\linewidth]{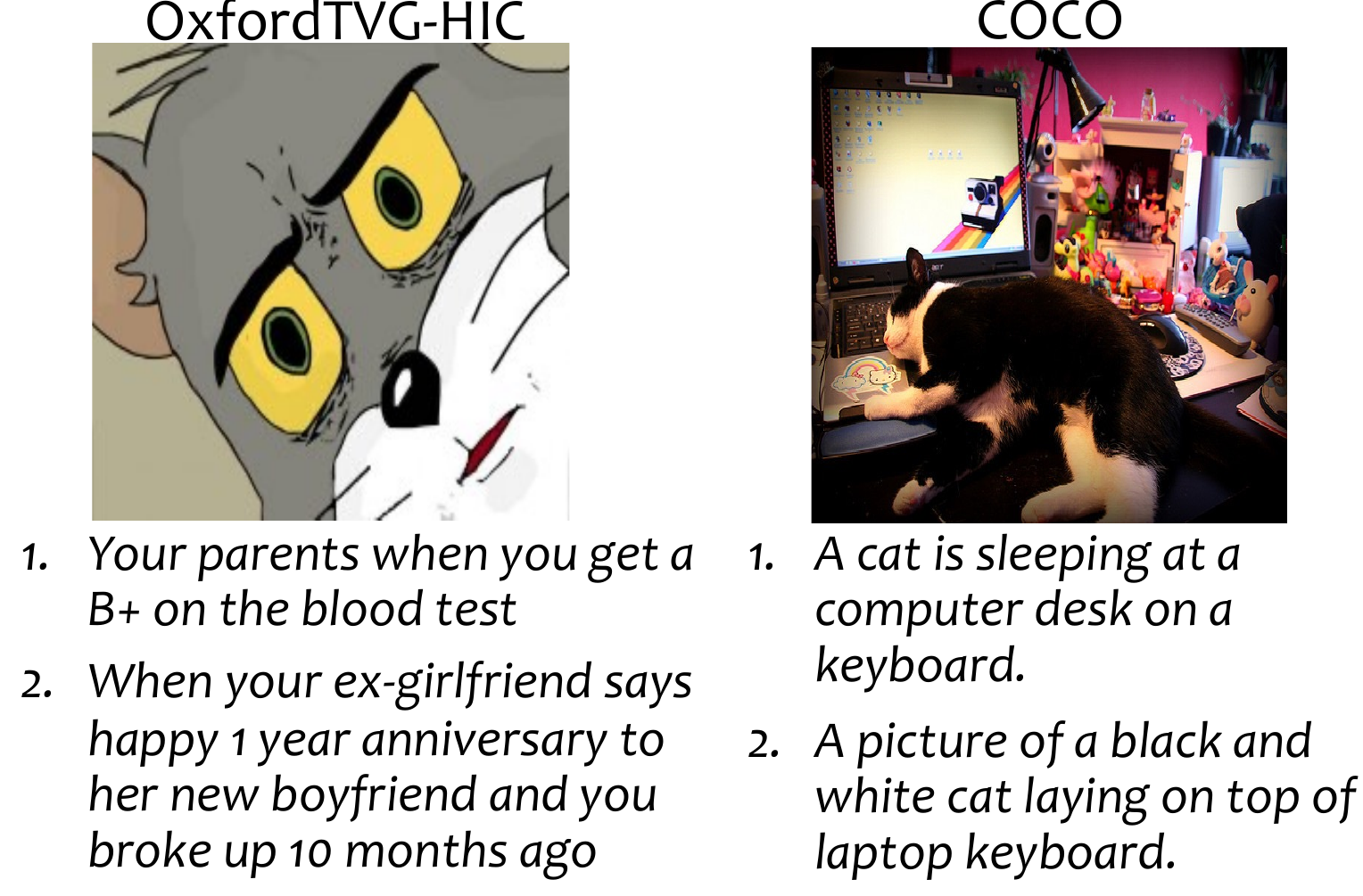}
\centering
}
\caption{\textbf{Image-text samples from \dname~and COCO}~\cite{mscoco}. In \dname~, captions for a cat image do not describe the physical features of the cat, but rather the situations that could elicit the cat’s facial expression. These situations create a humorous effect with the expression, as they are not offensive and violate the audience’s everyday-life expectations (Benign violation theory~\cite{benign_violation}). On the other hand, captions for a similar cat image in COCO~\cite{mscoco} explicitly describe the facts in the image. 
}
\label{fig:dataset_demo}
\end{figure}

%% file: tables/table_caption_dataset_statistics.tex
\begin{table}[t]
\centering
\tablestyle{1.6pt}{1}
\begin{tabular}{l|rrrc}

\Xhline{1.5pt}
Dataset & \multicolumn{1}{l}{\#Images} & \multicolumn{1}{l}{\#Captions} & \multicolumn{1}{l}{\makecell[c]{\#Captions\\/image}} & \makecell[c]{Dataset\\type} \\ \hline
\dname            & 54k                                        & 2885k                                       & \textbf{53.7}                                                             & Humour                  \\
COCO~\cite{mscoco}             & 123k                                       & 617k                                         & 5                                                                & Object                 \\
CC3M~\cite{conceptual}       & 3334k                                     & 3334k                                       & 1                                                                & Object                 \\
Flicker30k~\cite{flicker}          & 32k                                        & 159k                                         & 5                                                                & Object    \\
ArtEmis~\cite{artemis}   & 80k                                        & 455k                                         & 5.7                                                                & Emotion \\

ArtELingo~\cite{artelingo}   & 80k                                        & 1224k                                         & 15.3                                                                & Emotion \\
\Xhline{1.5pt}
\end{tabular}

\caption{\textbf{Image captioning dataset statistics comparison.} The average captions per image of \dname~is significantly higher than other captioning datasets, and the total number of captions is larger than most of the datasets.} \label{tab:caption_dataset_statistics}
\end{table}

%% file: figures/data_stats.tex
\begin{figure}[t]

\includegraphics[width=\linewidth]{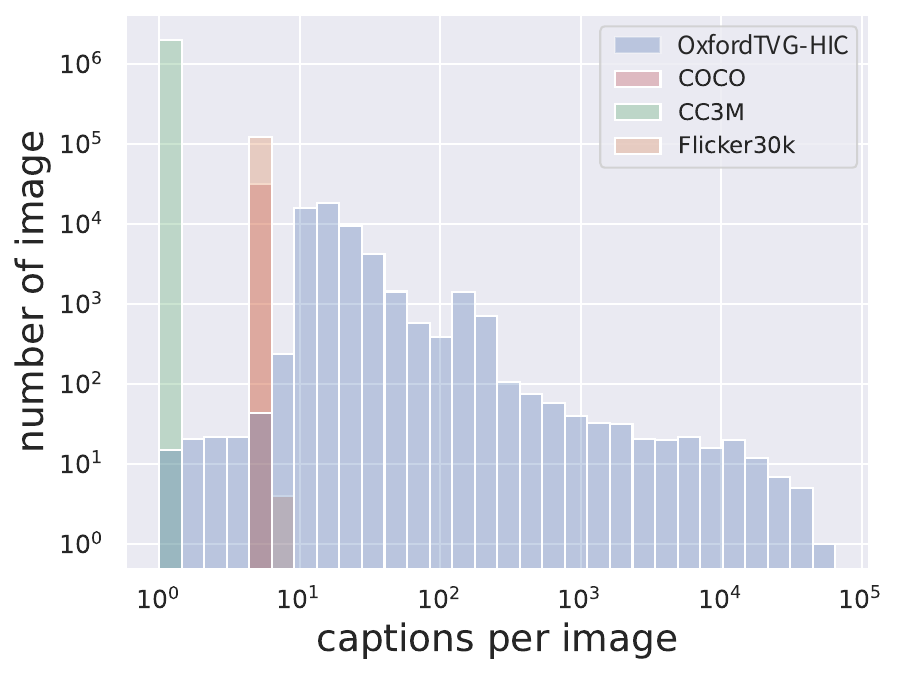}
\centering

\caption{
\dname~are much greater than the other image captioning datasets~\cite{mscoco, conceptual, flicker} in terms of the mean and variance for \textbf{the number of captions per image}. 
}
\label{fig:dataset_statistics}
\end{figure}

%% file: tables/table_emotion_div.tex
\begin{table}[t]
\tablestyle{1pt}{1.08}
\begin{tabular}{l|ccccccc}
\Xhline{1.5pt}
Dataset & \multicolumn{1}{c}{Anger} & \multicolumn{1}{c}{Digust} & \multicolumn{1}{c}{Fear} & \multicolumn{1}{c}{Joy} & \multicolumn{1}{c}{Neutral} & \multicolumn{1}{c}{Sad} & \multicolumn{1}{c}{Surprise} \\ \hline
Ours & \markcell{0.136} & \markcell{0.138} & \cellcolor{Gray!36} 0.058 & \cellcolor{Gray!36} 0.086 & \cellcolor{Gray!36} 0.363 & \markcell{0.078} & \markcell{0.145} \\
COCO\cite{mscoco} & 0.036 & 0.076 & \textbf{0.149} & 0.048 & \textbf{0.609} & 0.016 & 0.065 \\
CC3M\cite{conceptual} & 0.061 & 0.074 & 0.074 & \textbf{0.163} & 0.533 & 0.059 & 0.035 \\
Flicker30k\cite{flicker} & 0.049 & 0.136 & 0.129 & 0.074 & 0.549 & 0.023 & 0.037 \\ 
\Xhline{1.5pt}
\end{tabular}
\caption{\textbf{Emotion analysis for captions in \dname~and object-centric datasets.} The emotion richness of \dname~is at most $24.6\%$ greater than object-centric datasets.}
\label{tab:emotion_distribution}
\end{table}

%% file: tables/table_grammar.tex
\begin{table}[t]
\tablestyle{1.6pt}{1}
\centering
\begin{tabular}{l|c|c|c}
\Xhline{1.5pt}
\multirow{2}{*}{Dataset} & \multicolumn{3}{c}{\makecell[c]{\#Combination of Parts Of Speech}} \\
 & \multicolumn{1}{l|}{\makecell[c]{noun/verb}} & \multicolumn{1}{l|}{\makecell[c]{noun/verb/adj.}} & \multicolumn{1}{l}{\makecell[c]{noun/verb/adj./conj.}} \\ \hline
\dname            & \markcell{294,794}                                              & \markcell{455,280}                                                   & \markcell{545,013}                                                         \\
COCO~\cite{mscoco}             & 6,012                                                & 21,020                                                    & 33,997                                                          \\
CC3M~\cite{conceptual}       & 59,914                                              & 168,982                                                   & 225,933                                                         \\
Flicker30k~\cite{flicker}          & 11,796                                               & 34,034                                                    & 43,601                                                          \\
ArtEmis~\cite{artemis}          & 11,327                                               & 99,101                                                    & 153,324                                                          \\
ArtELingo~\cite{artelingo}          & 37,253                                               & 329,852                                                   & 452,531                                                          \\
\Xhline{1.5pt}
\end{tabular}
\caption{\textbf{Number of grammar patterns of image captioning datasets.} The grammar pattern is defined by the combination of the relative position of different POS (Parts Of Speech) in sentences. \dname~has a much larger number of grammar patterns than other captioning datasets.}
\label{tab:grammar_diversity}
\end{table}

%% file: figures/figure_data_semantic_dist.tex
\begin{figure}[t]
\resizebox{\columnwidth}{!}{%
\includegraphics[width=8cm]{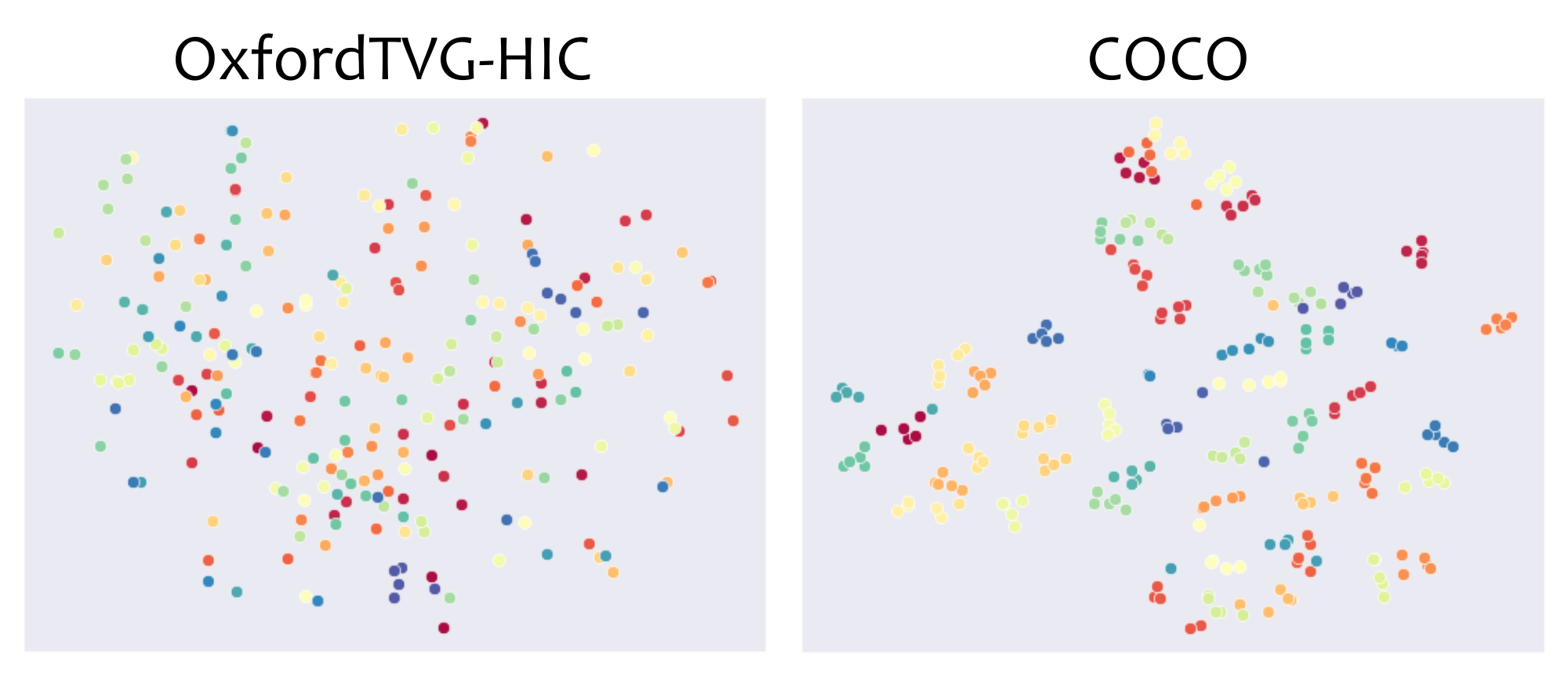}
\centering
}
\caption{\textbf{t-SNE distribution of sampled captions.} Points of the same colour denote captions of the same image. Captions of the same image are spread \textbf{distinct} in \dname, but they are clustered together in COCO~\cite{mscoco}.}
\label{fig:data_set_semantics}
\end{figure}

%% file: figures/figure_position.tex
\begin{figure}
    \resizebox{\columnwidth}{!}{%
        \centering
        \includegraphics[width=0.55\textwidth]{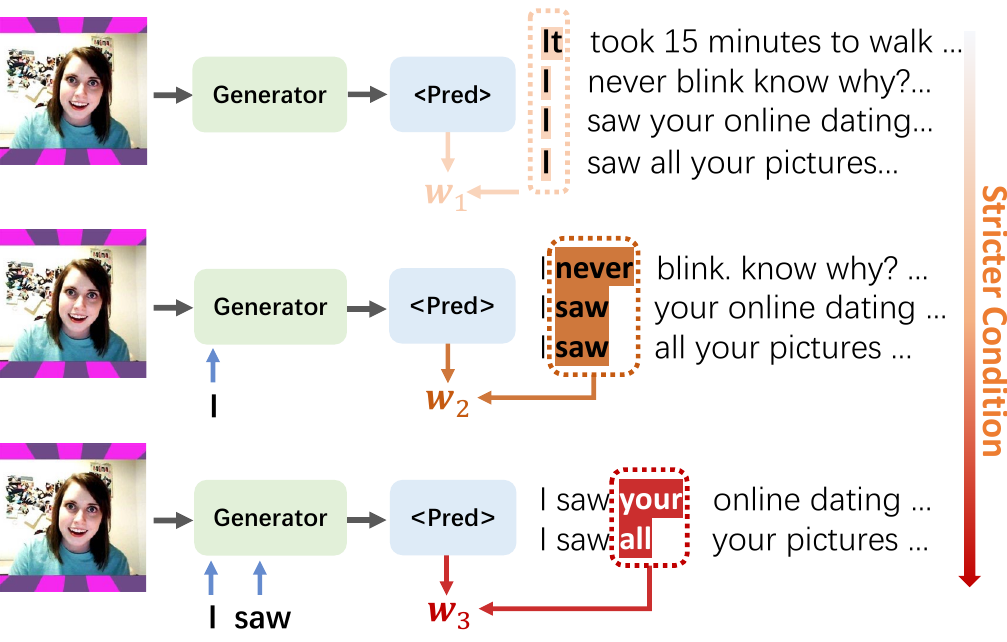}
    }
    \caption{\textbf{Intuition of the position-conditioned loss: stricter previous-word conditions constrain the next-word distribution.} For the first few tokens, there are diverse ground truths that exclude one another (more diverse than the demo here). In this case, we don't want to penalize false positive predictions too much because they are potentially ground truths. As more previous tokens are determined, the diversity of plausible ground truths of the next token decreases. Now the false positive predictions become less likely to be ground truths. Thus we increase the false positive loss weight $w_j$ to penalize those predictions. }
    \label{fig:position}
\end{figure}

%% file: figures/figure_generator_demo.tex
\begin{figure*}[t]
\tablestyle{1pt}{0.5}
\scalebox{0.9}{
\begin{tabular}[t]{lllll}
\includegraphics[width=0.25\textwidth,height=0.25\textwidth]{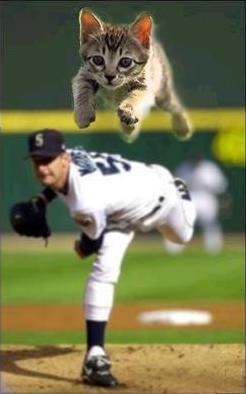} & \includegraphics[width=0.25\textwidth,height=0.25\textwidth]{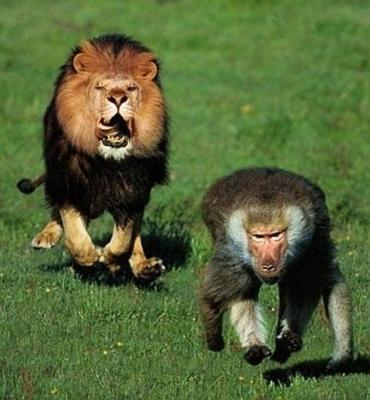} & \includegraphics[width=0.25\textwidth,height=0.25\textwidth]{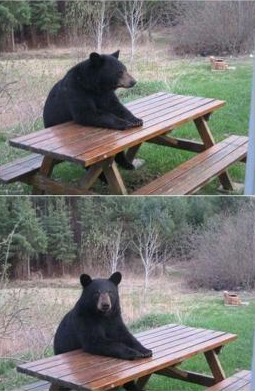} & \includegraphics[width=0.25\textwidth,height=0.25\textwidth]{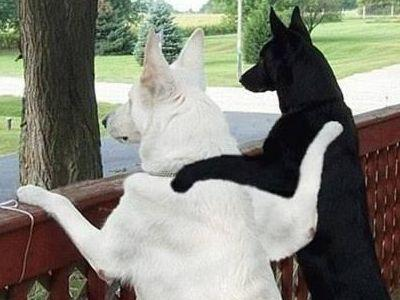} \\
 \makecell[tl]{I'm going to go to the bathroom}
 & \makecell[tl]{I'm going to go to the\\ convenience store}
 & \makecell[tl]{I'm going to give you a couple\\ of examples.}
 &Hey, hey, what's wrong with you?
\\
&&&& \\

\textcolor{RoyalBlue}{\makecell[tl]{Go Pikachu!}}
& \textcolor{RoyalBlue}{\makecell[tl]{Pool over! Give me your\\ driver's license.}}
& \textcolor{RoyalBlue}{\makecell[tl]{Next interview, please}} & \textcolor{RoyalBlue}{\makecell[tl]{The cats all over the country\\ have been deported as promised...}} \\
\end{tabular}
}
\caption{\textbf{Captions generated on unseen images by models that are trained on different losses}. Captions generated by the position-conditioned loss are rendered in \textcolor{RoyalBlue}{blue}, from which we can find that the position-conditioned loss solves the problem of limited diversity in cross-entropy. More examples will be shown in the appendix. }
\label{fig:generator_demo}
\end{figure*}

%% file: tables/table_humor_level.tex
\begin{table}[t]
\tablestyle{0.1pt}{1}
\centering
\begin{tabular}{cccccc}
\Xhline{1.5pt}
Model & Dataset & \textit{w/} \makecell[c]{position-\\condition?}            & \makecell[c]{Humour    \\ score$\uparrow$}     & \makecell[c]{Benign \\ score~\cite{villo}$\uparrow$}  & \makecell[c]{Fluency \\ (Parrot~\cite{parrot})$\uparrow$} \\ \hline
\multirow{3}{*}{\makecell[c]{BLIP~\\\cite{li2022blip}}} & COCO~\cite{mscoco}             & $\times$              & \multicolumn{1}{c}{0.416} & \textbf{0.999} & \textbf{0.956} \\
        & Ours            & $\times$              & \multicolumn{1}{c}{0.749} & 0.990 & 0.897 \\
        & Ours            & \checkmark & \markcell{0.828} & \cellcolor{Gray!36}0.991 & \cellcolor{Gray!36}0.899 \\
       \hline
\multirow{3}{*}{\makecell[c]{ClipCap~\\\cite{clipcap}}} & COCO~\cite{mscoco}             & $\times$              & \multicolumn{1}{c}{0.270} & \textbf{1.000} & 0.899 \\
        & Ours            & $\times$              & \multicolumn{1}{c}{0.650} & 0.935 & \textbf{0.901} \\
        & Ours            & \checkmark & \markcell{0.832} & \cellcolor{Gray!36}0.984 & \cellcolor{Gray!36}0.831 \\
    \Xhline{1.5pt}
\end{tabular}
\caption{\textbf{Evaluation results of humour generators}. Humour scores of models trained on \dname~and position-conditioned loss are the highest as expected while the benign and fluency scores for all listed models are comparable, which indicates models trained on \dname~with position-conditioned loss generate the highest humour intensity.} \label{tab:humor level}
\end{table}

%% file: tables/table_kernel_function.tex
\begin{table}[t]
\tablestyle{1.6pt}{1}
\centering
\begin{tabular}{c|ccccc}
\Xhline{1.5pt}
\makecell[c]{Kernel \\function}  & \makecell[c]{Humour\\score$\uparrow$}         & \makecell[c]{Benign \\ score~\cite{villo}$\uparrow$}  & \makecell[c]{Fluency \\ (Parrot~\cite{parrot})$\uparrow$}   & \makecell[c]{Diversity\\(CLIP~\cite{clip}))$\uparrow$}              \\ \hline
sigmoid                                       &\markcell{0.832} & \markcell{0.984} &\cellcolor{Gray!36} 0.831 &\markcell{0.373} \\
  
linear                                           &      0.792      & 0.911&             0.795         &             0.348                     \\
 
gaussian                                            &     0.826     & 0.906  &             \textbf{0.873}           &  0.362                                 \\ \Xhline{1.5pt}                 
\end{tabular}
\caption{\textbf{Influence of kernel functions of position-conditioned loss on the performance of ClipCap}~\cite{clipcap}. Experimentally, the sigmoid kernel function conveys the highest benign and humour scores, or equivalently, the highest humour intensity. Besides, models trained with the sigmoid kernel loss generate the most diverse captions.}\label{tab:kernel_function}
\end{table}

%% file: tables/table_ablation.tex
\begin{table}[t]
\tablestyle{1.6pt}{1}
\centering
\begin{tabular}{ccccc}
\Xhline{1.5pt}
 \multicolumn{1}{l}{$\alpha$} & \makecell[c]{Humour\\score$\uparrow$}      & \makecell[c]{Benign \\ score~\cite{villo}$\uparrow$}  & \makecell[c]{Fluency \\ (Parrot~\cite{parrot})$\uparrow$}         & \makecell[c]{Diversity\\(CLIP~\cite{clip}))$\uparrow$}              \\ \hline
 / & 0.650& 0.935 & \textbf{0.901} & 0.254\\
 2                                  & \multicolumn{1}{c}{0.818} & 0.909 & 0.881 &\multicolumn{1}{c}{0.358} \\
         4                                & \multicolumn{1}{c}{0.823} & 0.911& 0.889 &\multicolumn{1}{c}{0.362} \\
         6                                  & \markcell{0.832} & \markcell{0.984} &\cellcolor{Gray!36} 0.831 &\cellcolor{Gray!36}0.373 \\
        8                                  & \multicolumn{1}{c}{0.822} & 0.905& 0.795 &\textbf{0.384} \\
 \Xhline{1.5pt}                 
\end{tabular}
\caption{\textbf{Influence of the hyper-parameter $\alpha$ of the sigmoid kernel function defined in Eq. \eqref{eq:sigmoid} on diversity and humour level of ClipCap}~\cite{clipcap}. As $\alpha$ increases, the false positive penalty increases slowlier and more diversity is introduced because the supervision is less strict. Also, models trained on position-conditioned loss generally tend to have more diversity than cross-entropy (denoted by $\alpha=$``/").} \label{tab:ablation}
\end{table}

%% file: figures/figure_attentionmap_demo.tex
\begin{figure*}[h]
\tablestyle{0.3pt}{0.2}
\scalebox{0.88}{
\begin{tabular}{cccccc}
COCO & Ours & COCO & Ours & COCO & Ours \\ 
    \includegraphics[width=0.16\textwidth,height=0.16\textwidth]{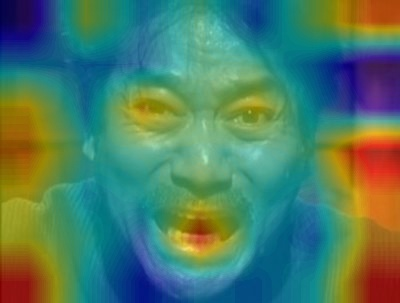} & \includegraphics[width=0.16\textwidth,height=0.16\textwidth]{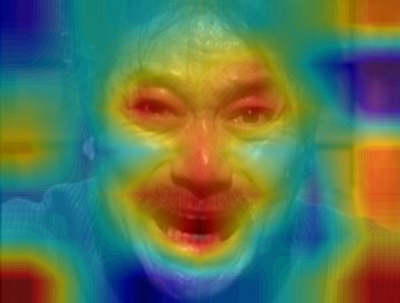}& \includegraphics[width=0.16\textwidth,height=0.16\textwidth]{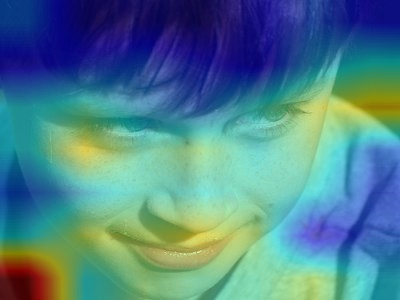}& \includegraphics[width=0.16\textwidth,height=0.16\textwidth]{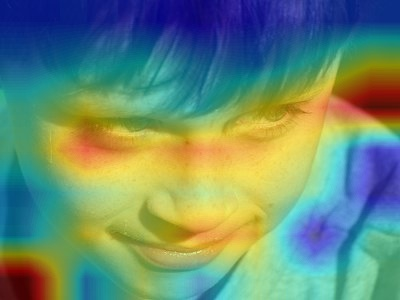}& \includegraphics[width=0.16\textwidth,height=0.16\textwidth]{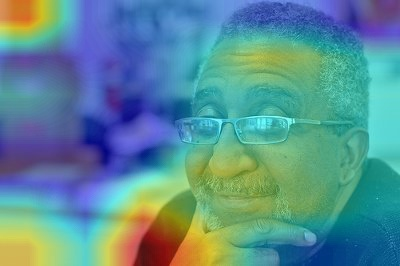}& \includegraphics[width=0.16\textwidth,height=0.16\textwidth]{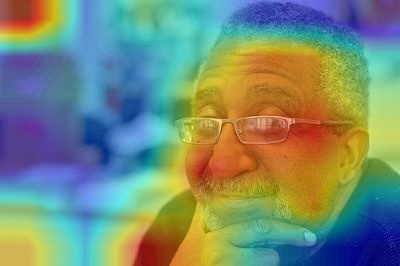}\\
     \includegraphics[width=0.16\textwidth,height=0.16\textwidth]{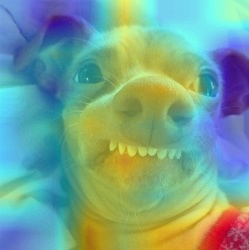} & \includegraphics[width=0.16\textwidth,height=0.16\textwidth]{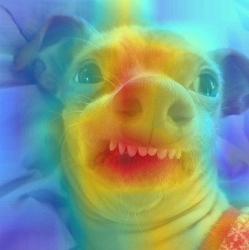}& \includegraphics[width=0.16\textwidth,height=0.16\textwidth]{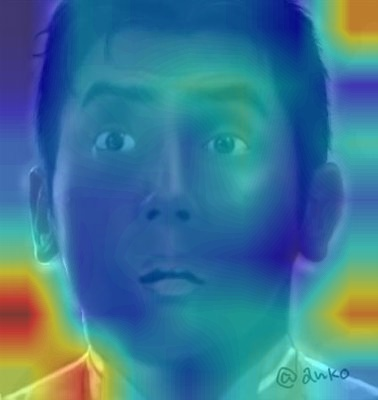}& \includegraphics[width=0.16\textwidth,height=0.16\textwidth]{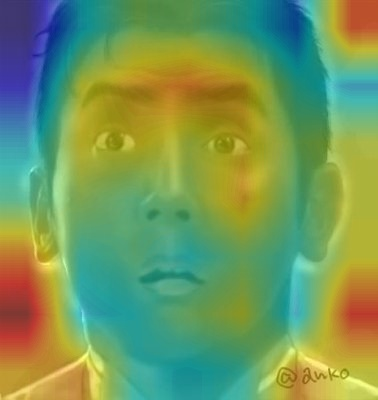}& \includegraphics[width=0.16\textwidth,height=0.16\textwidth]{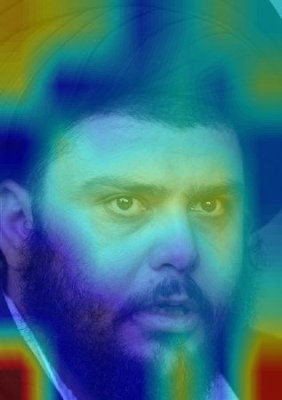}& \includegraphics[width=0.16\textwidth,height=0.16\textwidth]{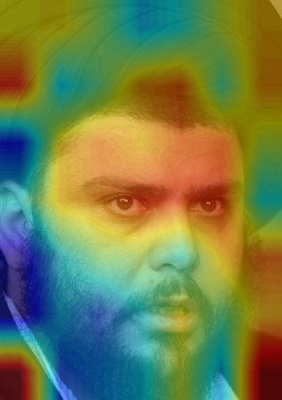}\\
\end{tabular}
}
\caption{\textbf{Attention heatmap of the humour generator image encoder.} The generator tends to focus more on facial expressions, especially the ``dramatic" parts which involve more emotions.}
\label{fig:attention_demo}
\end{figure*}

%% file: figures/figure_grad_cam_demo.tex
\begin{figure*}[h]

\includegraphics[width=0.88\textwidth]{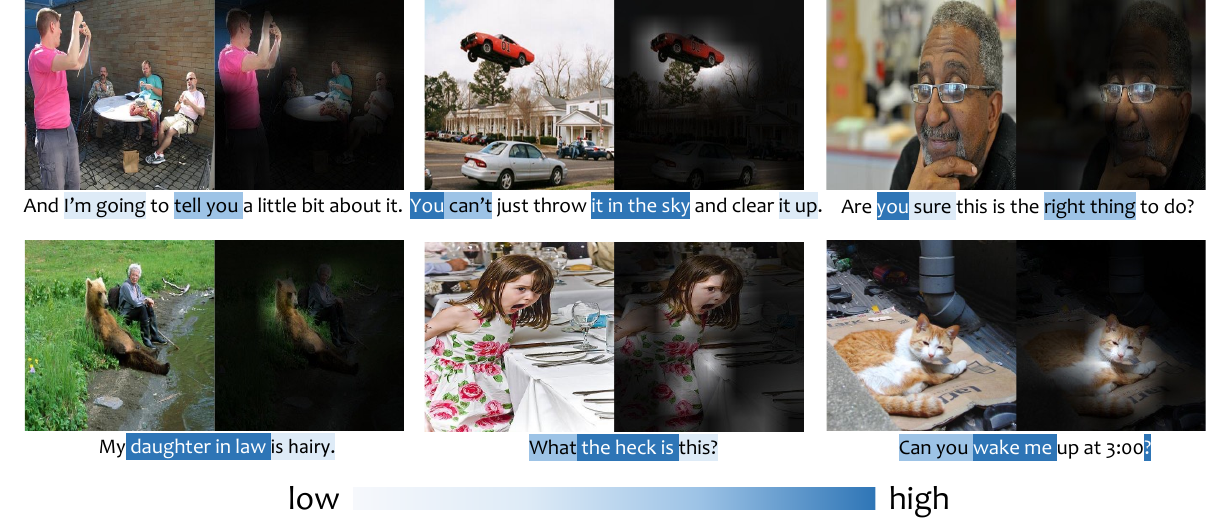}
\centering

\caption{\textbf{Gradient visualization of image-text pairs from humour classifier.} The evaluator tends to focus on abnormal parts and facial expressions of the images  which involve more emotion; More attention is paid to the pronouns in the text which induces engagement from the audience. Deeper colour on the texts means a larger contribution to humour.}\label{fig:grad_cam_demo}
\end{figure*}

%% file: supp/table/table_hic_demo.tex
\begin{figure*}[t]
\tablestyle{0.1pt}{0.5}
\centering
\scalebox{0.85}{
\begin{tabular}[t]{llllll}
\includegraphics[width=0.23\textwidth,height=0.23\textwidth]{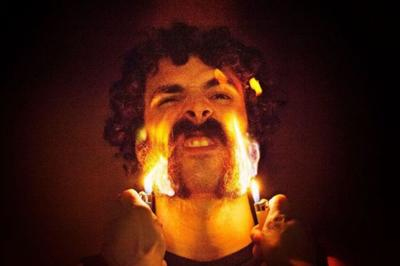} & \includegraphics[width=0.23\textwidth,height=0.23\textwidth]{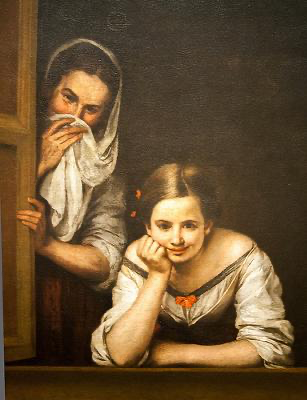} & \includegraphics[width=0.23\textwidth,height=0.23\textwidth]{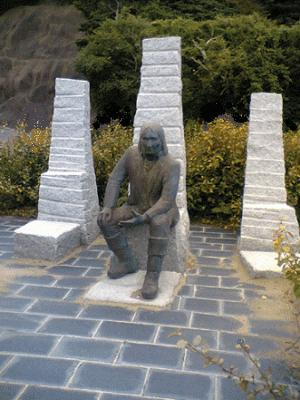} & \includegraphics[width=0.23\textwidth,height=0.23\textwidth]{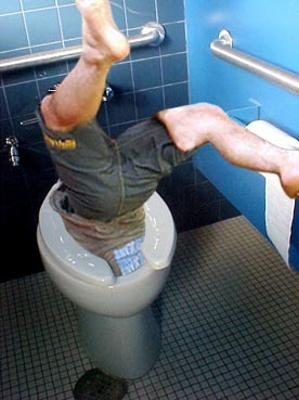} & \includegraphics[width=0.23\textwidth,height=0.23\textwidth]{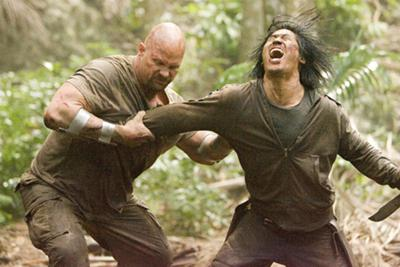}
\\
&&&&& \\

\makecell[tc]{I hate shaving because \\it hurts so much.}
& \makecell[tc]{The smell of my fart \\stinks more}
& \makecell[tc]{I'm sitting here and playing\\ with a nice guy called a stone} 

& \makecell[tc]{Second entrance to\\ four-dimensional space.}
& \makecell[tc]{Go back to school!}
\\
&&&&& \\
\includegraphics[width=0.23\textwidth,height=0.23\textwidth]{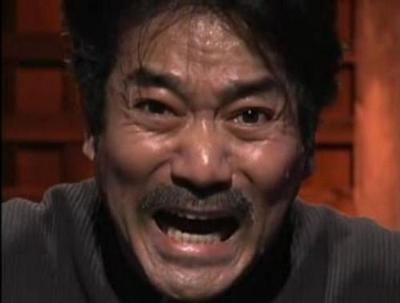} & \includegraphics[width=0.23\textwidth,height=0.23\textwidth]{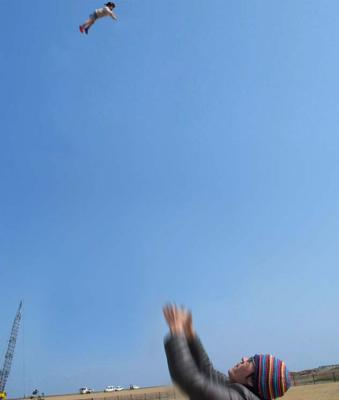} & \includegraphics[width=0.23\textwidth,height=0.23\textwidth]{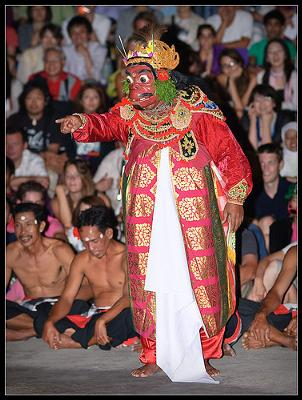}& \includegraphics[width=0.23\textwidth,height=0.23\textwidth]{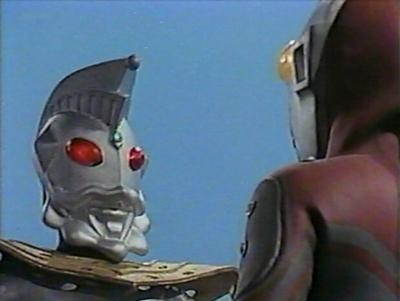} & \includegraphics[width=0.23\textwidth,height=0.23\textwidth]{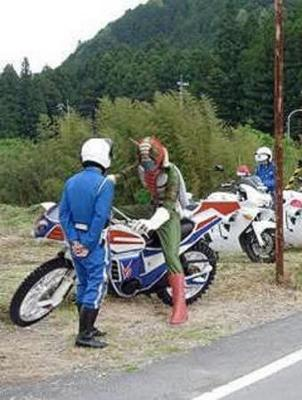}
\\
&&&&& \\

\makecell[tc]{The expression on the face\\ of the next patient.}
& \makecell[tc]{I was up in the air,\\ but I didn't fly.}
& \makecell[tc]{I'm the head of the\\ Strawberries team.} 
& \makecell[tc]{Your eyes are red...\\ are you angry?}
& \makecell[tc]{"You're fighting for the world." }\\

&&&&& \\
\includegraphics[width=0.23\textwidth,height=0.23\textwidth]{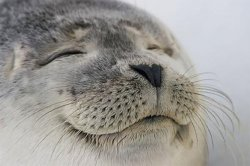} & \includegraphics[width=0.23\textwidth,height=0.23\textwidth]{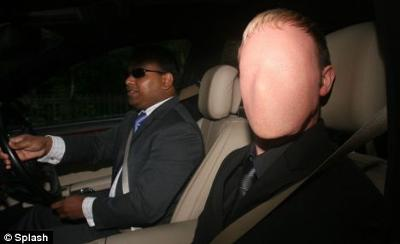} & \includegraphics[width=0.23\textwidth,height=0.23\textwidth]{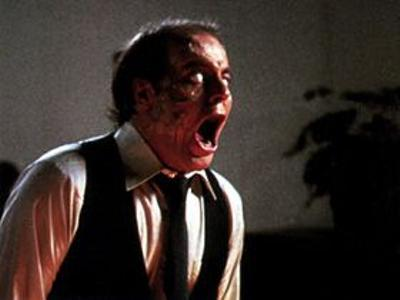}& \includegraphics[width=0.23\textwidth,height=0.23\textwidth]{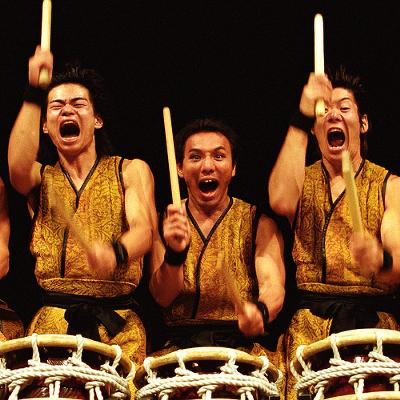} & \includegraphics[width=0.23\textwidth,height=0.23\textwidth]{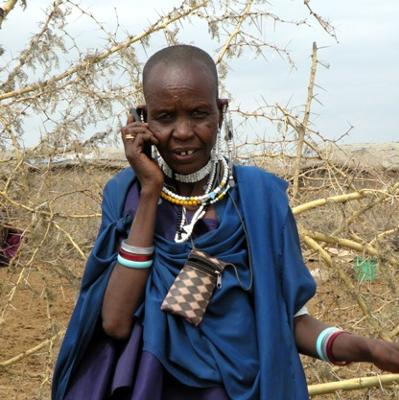}
\\
&&&&& \\

\makecell[tc]{Awkward moment when you\\leave a store without buying}
& \makecell[tc]{Watch my eyes.}
& \makecell[tc]{When your boyfriend \\knows you are pregnant.} 
& \makecell[tc]{The front row of the\\ roller coaster.} 
& \makecell[tc]{What are you talking about?\\ What are YOU talking ABOUT?}\\

&&&&& \\
\includegraphics[width=0.23\textwidth,height=0.23\textwidth]{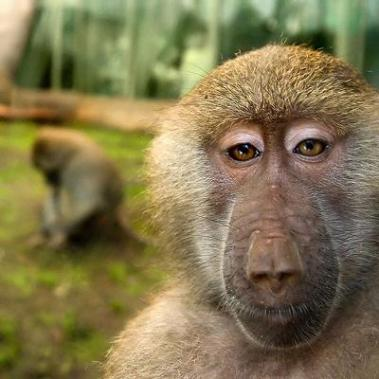} & \includegraphics[width=0.23\textwidth,height=0.23\textwidth]{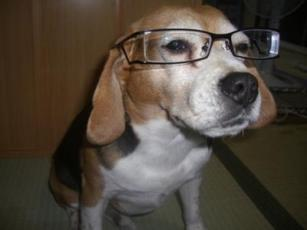} & \includegraphics[width=0.23\textwidth,height=0.23\textwidth]{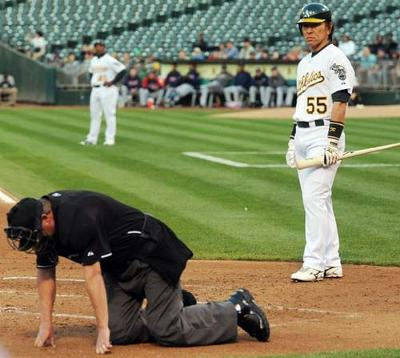} & \includegraphics[width=0.23\textwidth,height=0.23\textwidth]{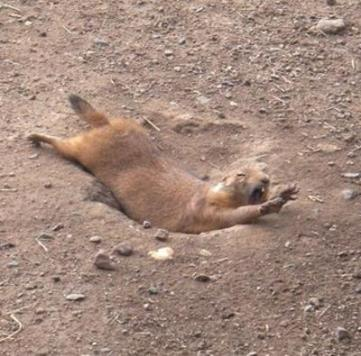}& \includegraphics[width=0.23\textwidth,height=0.23\textwidth]{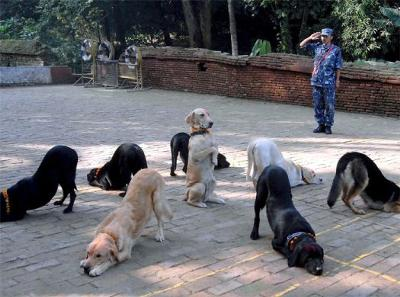}
\\
&&&&& \\

\makecell[tc]{My husband opens the window\\ and looks at his son in the park}
& \makecell[tc]{I'm a dog, right?}
& \makecell[tc]{The scene of a wild monkey\\ being dragged away by a turtce. } 
& \makecell[tc]{I jumped into the pool,\\ and there was no water. } 
& \makecell[tc]{"Hey, guys, \\let's get out of here!"}\\

&&&&& \\
\includegraphics[width=0.23\textwidth,height=0.23\textwidth]{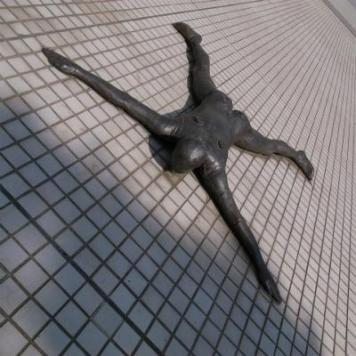} & \includegraphics[width=0.23\textwidth,height=0.23\textwidth]{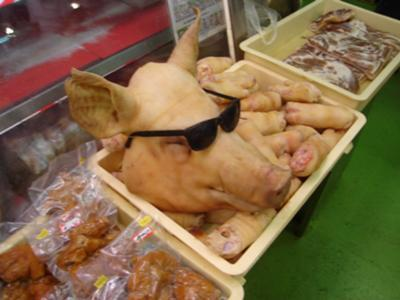} & \includegraphics[width=0.23\textwidth,height=0.23\textwidth]{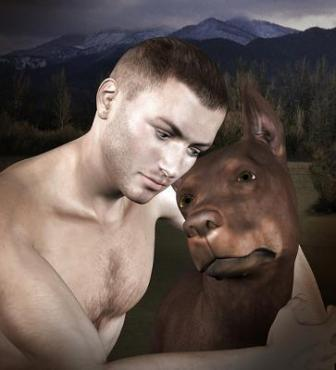} & \includegraphics[width=0.23\textwidth,height=0.23\textwidth]{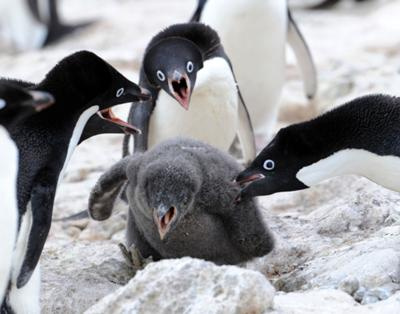}& \includegraphics[width=0.23\textwidth,height=0.23\textwidth]{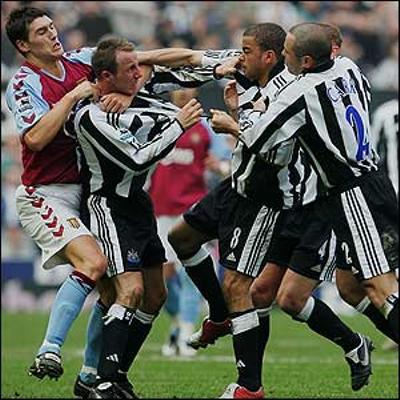}
\\
&&&&& \\

\makecell[tc]{I've been frogs\\for a long time.}
& \makecell[tc]{You've never been \\a pig before, have you?}
& \makecell[tc]{I remember when your\\husband was a cow. } 
& \makecell[tc]{My girlfriend \\when I refused to confess \\I was surrounded by girls.} 
& \makecell[tc]{You're mine!}\\
\end{tabular}
}
\caption{\textbf{Humorous captions generated on unseen images in the \dname~dataset.} The humour generators have been able to generate funny captions given a wide variety of unseen images including paintings, animation, scenes of nature, animals, and human expression. Thus, the humour generators exhibit promising generalisability.}
\label{fig:hic_generated_demo}
\end{figure*}

%% file: supp/table/table_coco_demo.tex
\begin{figure*}[t]
\tablestyle{0.11pt}{0.5}
\centering
\scalebox{0.85}{
\begin{tabular}[t]{llllll}
\includegraphics[width=0.23\textwidth,height=0.23\textwidth]{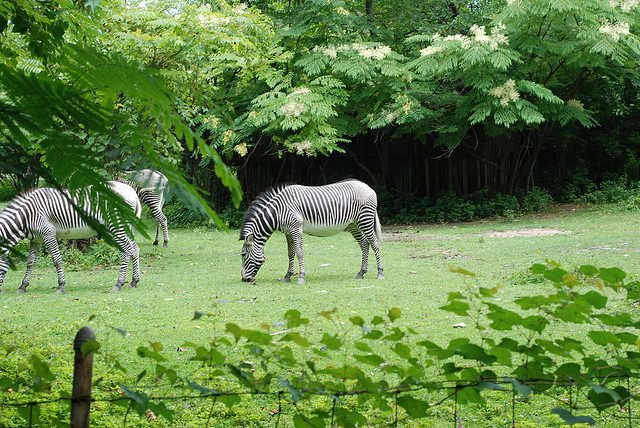} & \includegraphics[width=0.23\textwidth,height=0.23\textwidth]{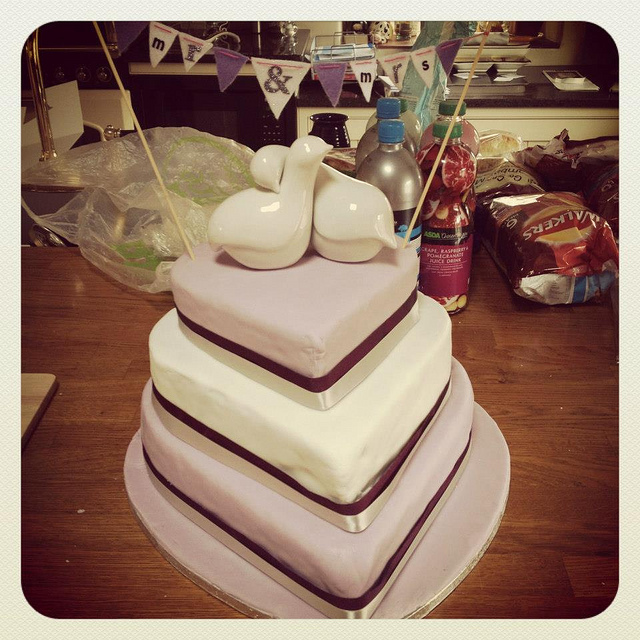} & \includegraphics[width=0.23\textwidth,height=0.23\textwidth]{supp/coco_funny_imgs/demo3.png} & \includegraphics[width=0.23\textwidth,height=0.23\textwidth]{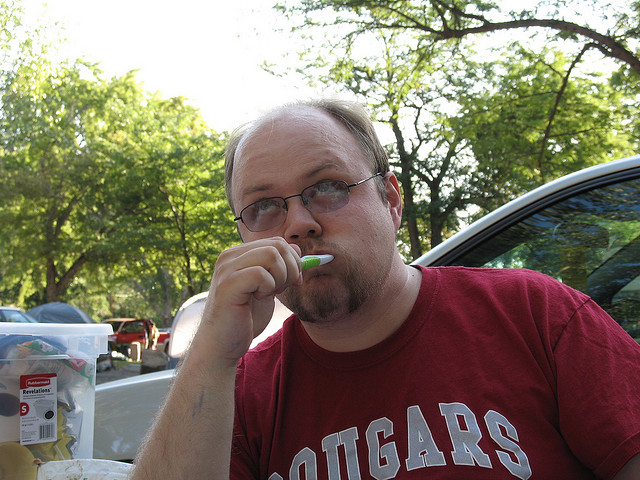} & \includegraphics[width=0.23\textwidth,height=0.23\textwidth]{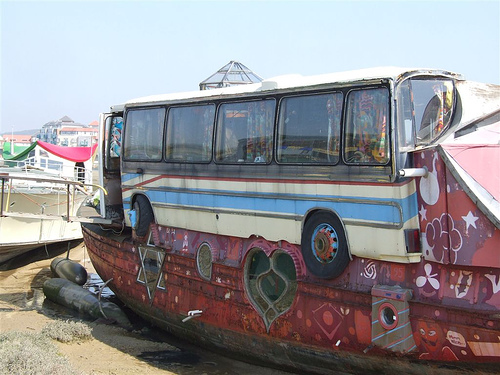}
\\
&&&&& \\

\makecell[tc]{Oreo horses are more \\expensive than horses.}
& \makecell[tc]{Let's stay on the honeymoon.}
& \makecell[tc]{How many years did\\ you drink water?} 

& \makecell[tc]{I'm not fat, I'm \\just a little bumpy.}
& \makecell[tc]{US marine corps,\\ America's best organization.}

\\
&&&&& \\
\includegraphics[width=0.23\textwidth,height=0.23\textwidth]{supp/coco_funny_imgs/demo6.png} & \includegraphics[width=0.23\textwidth,height=0.23\textwidth]{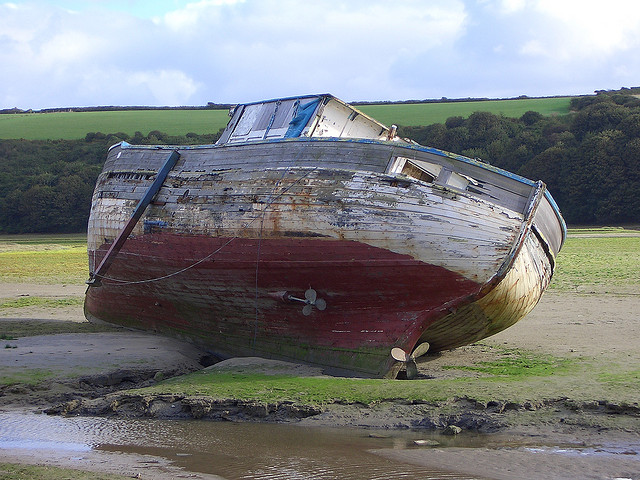} & \includegraphics[width=0.23\textwidth,height=0.23\textwidth]{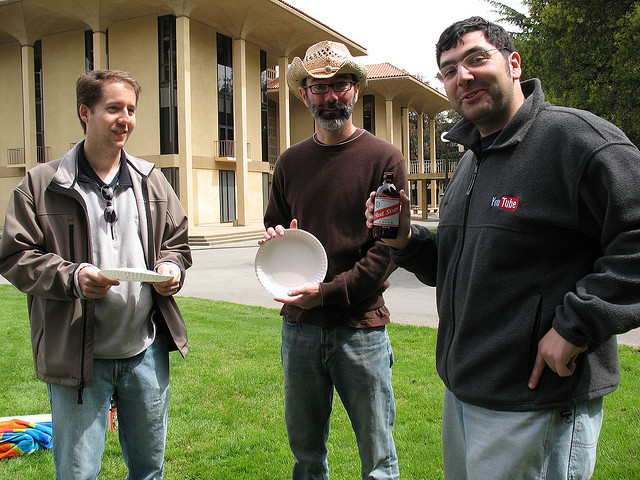}& \includegraphics[width=0.23\textwidth,height=0.23\textwidth]{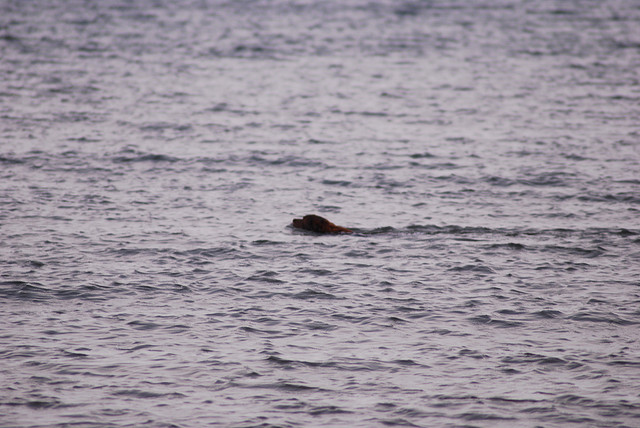} & \includegraphics[width=0.23\textwidth,height=0.23\textwidth]{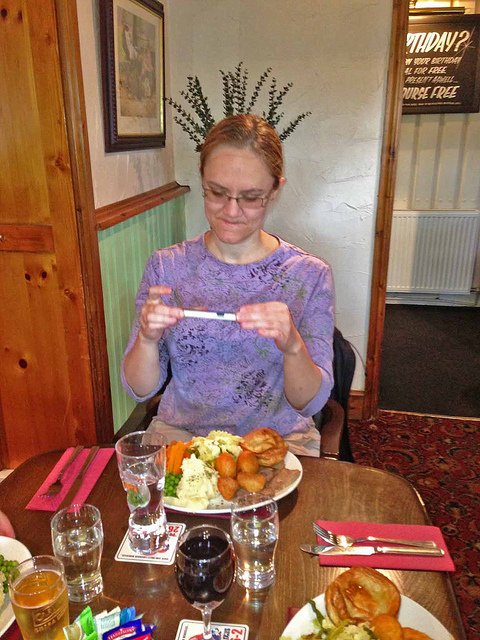}
\\
&&&&& \\

\makecell[tc]{The claws of the enemy are\\ not aware of my suffering.}
& \makecell[tc]{I can't get out. \\It's starting to freak me out}
& \makecell[tc]{Hi, this is the \\urine bottle for today. } 
& \makecell[tc]{Don't fall in love,\\ fall in water.}
& \makecell[tc]{Do I have to pay for it?}\\

&&&&& \\
\includegraphics[width=0.23\textwidth,height=0.23\textwidth]{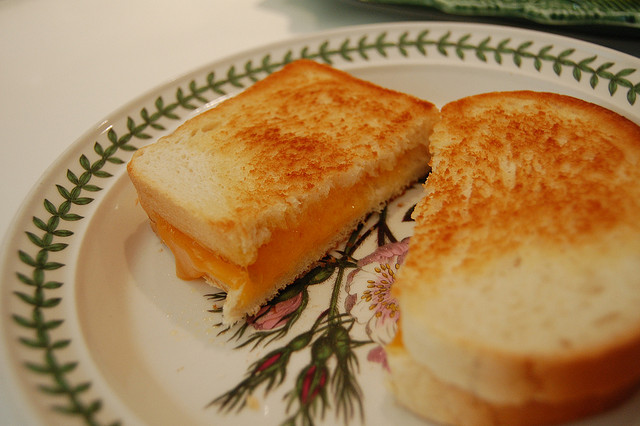} & \includegraphics[width=0.23\textwidth,height=0.23\textwidth]{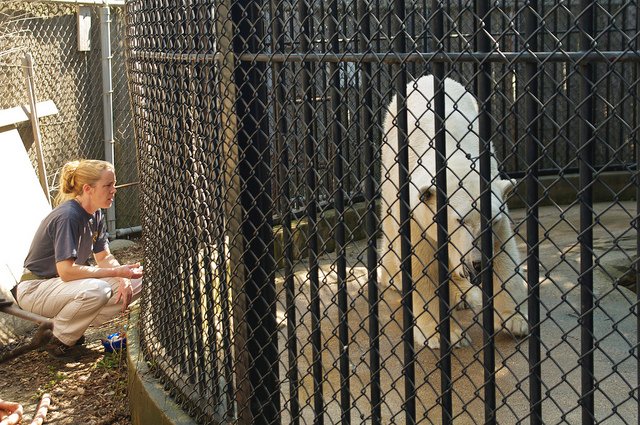} & \includegraphics[width=0.23\textwidth,height=0.23\textwidth]{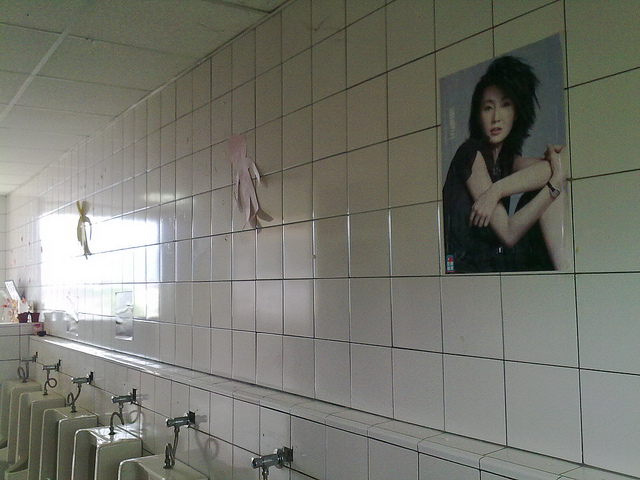}& \includegraphics[width=0.23\textwidth,height=0.23\textwidth]{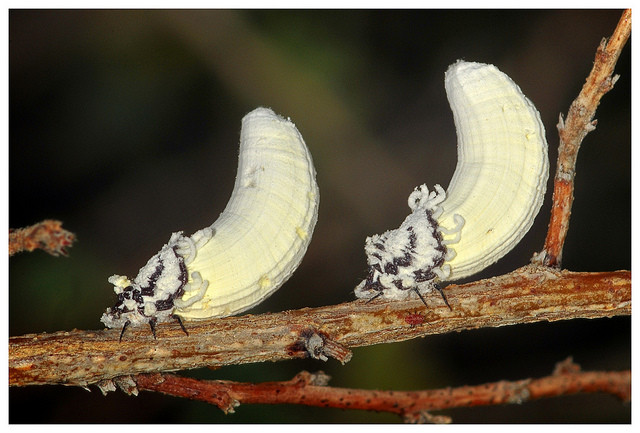} & \includegraphics[width=0.23\textwidth,height=0.23\textwidth]{supp/coco_funny_imgs/demo15.png}
\\
&&&&& \\

\makecell[tc]{The face of my daughter-in-law\\ was going to fall apart. }
& \makecell[tc]{I've been in jail for years.\\ I've never been a criminal.}
& \makecell[tc]{The wall is so hot. \\I can't get out. Help me.} 
& \makecell[tc]{Darling you smell \\like banana ice cream} 
& \makecell[tc]{I like your garden.}\\

\end{tabular}
}
\caption{\textbf{Humorous captions generated on unseen images in COCO}~\cite{mscoco}. The humour generators manage to generate funny captions of unseen images from other object-centric datasets such as COCO~\cite{mscoco} as well, which further demonstrates the generalisability of our trained humour generators and the diversity of \dname~in an implicit manner.}
\label{fig:coco_generated_demo}
\end{figure*}

%% file: supp/table/table_hic_failure.tex
\begin{figure*}[t]
\tablestyle{1pt}{0.5}
\scalebox{0.85}{
\begin{tabular}[t]{llllll}
\includegraphics[width=0.23\textwidth,height=0.23\textwidth]{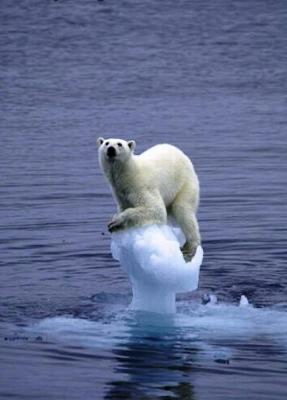} & \includegraphics[width=0.23\textwidth,height=0.23\textwidth]{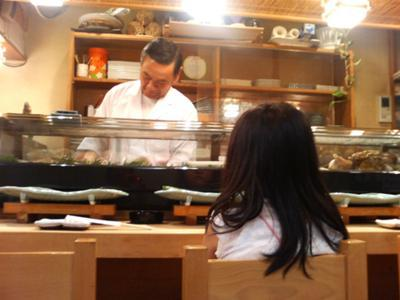} & \includegraphics[width=0.23\textwidth,height=0.23\textwidth]{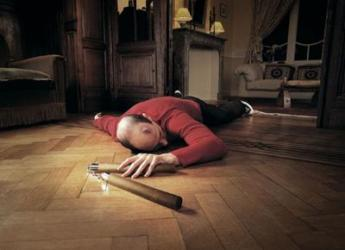} & \includegraphics[width=0.23\textwidth,height=0.23\textwidth]{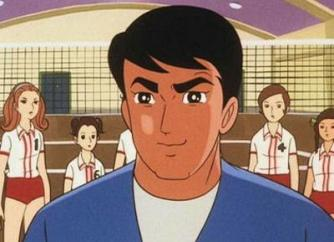} & \includegraphics[width=0.23\textwidth,height=0.23\textwidth]{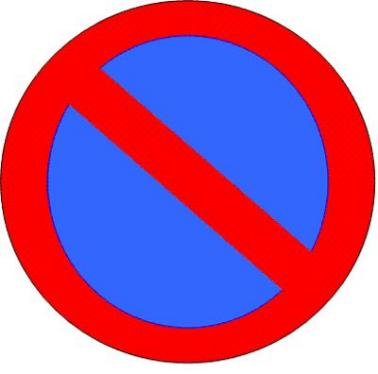}
\\
&&&&& \\

\makecell[tc]{Let’s sink}
& \makecell[tc]{This is a cheap restaurant.}
& \makecell[tc]{A game of death.} 

& \makecell[tc]{Please don't go there \\until you're a professional. }
& \makecell[tc]{You can't be in and out of \\the house in a stakeout. }
\\

\end{tabular}
}
\caption{\textbf{Failed captions generated on unseen images in \dname}. While the generations of our humour generators are novel and an initial attempt towards a new direction: humour-oriented captioning, they still require significant improvement to match the sophistication and nuance of their human-made counterparts. Here, we illustrate how our neural speakers fail to elicit the perception of humour even if the generated captions are somewhat relevant to the input images.}
\label{fig:hic_generated_failure_demo}
\end{figure*}